\documentclass{article}
\usepackage{multirow}

\usepackage[preprint]{neurips_2023}

\usepackage[utf8]{inputenc} 
\usepackage[T1]{fontenc}    
\usepackage[hidelinks,colorlinks=true]{hyperref}       
\usepackage{url}            
\usepackage{booktabs}       
\usepackage{amsfonts}       
\usepackage{nicefrac}       
\usepackage{microtype}      
\usepackage{xcolor}         

\usepackage{float}  
\usepackage{graphicx} 
\usepackage{caption}
\usepackage{subcaption}
\usepackage{booktabs}
\usepackage{amsmath}
\usepackage{amssymb}
\usepackage{xspace}
\usepackage{arydshln} 
\usepackage[misc]{ifsym}

\hypersetup{
    colorlinks=true,
    linkcolor=blue,    
    citecolor=blue,    
    urlcolor=blue      
}

\newcommand{\llamoe}{{LLaMA-MoE}\xspace}
\newcommand{\llama}{LLaMA\xspace}
\newcommand{\moe}{MoE\xspace}

\definecolor{color1}{rgb}{0.1,0.7,0.8}
\definecolor{color2}{rgb}{0.9,0.1,0.1}
\definecolor{color3}{rgb}{0.7,0.3,0.7}
\definecolor{color4}{rgb}{0.3,0.3,0.7}
\definecolor{color5}{RGB}{8, 102, 3}
\definecolor{color6}{rgb}{0.53, 0.66, 0.42}

\title{
LLaMA-MoE: Building Mixture-of-Experts \\ from LLaMA with Continual Pre-training
}

\author{
\textbf{Tong Zhu}$^{1}$\thanks{Work was done during an internship at Shanghai AI Laboratory.} , \textbf{Xiaoye Qu}$^{2}$, \textbf{Daize Dong}$^{2}$, \textbf{Jiacheng Ruan}$^{3}$, \textbf{Jingqi Tong}$^{4}$, \\ \textbf{Conghui He}$^{2}$, \textbf{Yu Cheng}$^{5 \text{ \Letter}}$ \\
$^{1}$ Soochow University $^{2}$ Shanghai AI Laboratory 
$^{3}$ Shanghai Jiao Tong University \\
$^{4}$ Fudan University $^{5}$ The Chinese University of Hong Kong \\
\texttt{tzhu7@stu.suda.edu.cn,} \texttt{\{quxiaoye,dongdaize.d,heconghui\}@pjlab.org.cn,} \\
\texttt{jackchenruan@sjtu.edu.cn,} \texttt{jqtong23@m.fudan.edu.cn,} \texttt{chengyu@cse.cuhk.edu.hk}
}

\begin{document}

\maketitle

\begin{abstract}

Mixture-of-Experts (MoE) has gained increasing popularity as a promising framework for scaling up large language models (LLMs). 
However, training MoE from scratch in a large-scale setting still suffers from data-hungry and instability problems.
Motivated by this limit, we investigate building MoE models from existing dense large language models.
Specifically, based on the well-known \llama-2 7B model, we obtain an MoE model by: (1) \textit{Expert Construction}, which partitions the parameters of original Feed-Forward Networks (FFNs) into multiple experts; (2) \textit{Continual Pre-training}, which further trains the transformed MoE model and additional gate networks. 
In this paper, we comprehensively explore different methods for expert construction and various data sampling strategies for continual pre-training.
After these stages, our \llamoe models could maintain language abilities and route the input tokens to specific experts with part of the parameters activated.
Empirically, by training 200B tokens, \llamoe-3.5B models significantly outperform dense models that contain similar activation parameters. 
The source codes and models are available at \url{https://github.com/pjlab-sys4nlp/llama-moe} .

\end{abstract}

\section{Introduction}

Large language models (LLMs) \citep{chatgpt,touvron2023llama,cai2024internlm2} have presented remarkable understanding and reasoning capabilities on a wide range of tasks~\citep{lu2024mitigating,su2024living,su2024timo,tao2024magis}.
Nowadays, scaling model size has become the de facto approach to further boost the overall performances~\citep{wei2022emergent,hoffmann2022training}.
However, the immense model size is unsustainable due to the computational costs~\citep{fedus2022review}.
Motivated by the trade-off between scaling the model size and reducing the inference cost, sparse Mixture-of-Experts (MoE) is further proposed to activate a part of the model parameters~\citep{shazeer2017outrageously,lepikhin2020gshard}.
Inspired by this, we focus on sparsely activated MoE models that decouple model size from computation costs.

However, training MoEs from scratch \citep{fedus2022switch,zoph2022st,xue2024openmoe,dai2024deepseekmoe} leads to a significant overall budget.
In this work, we reduce the training costs by investigating building MoE models from existing dense LLMs. 
Moreover, starting from the dense model provides flexible structure design choices for MoE.
In other words, we can place the MoE block in any transformer layers. 
In this paper, we dedicate to building a full MoE model from LLaMA~\citep{llama2}, where each layer contains an MoE block.

To build strong \llamoe models, we identify two important challenges. 
First, \textbf{how to effectively construct experts} from the Feed-Forward Networks (FFNs) in the existing LLMs. 
There are works exploring splitting FFN parameters to construct experts on T5~\citep{zhang2021moefication} or BERT~\citep{zuo2022moebert} models. 
Conversely, \citet{komatsuzaki2022sparse} directly copy the FFNs to form experts.
However, there is no existing work exploring it for decoder-only models. 
Notably, the Swish-based FFN~\citep{ramachandran2017swish,shazeer2020glu} does not bring natural sparsity as ReLU~\citep{nair2010relu} in T5 and BERT.
Second, \textbf{overcoming the performance decrease entailed by
changing the network structure} from dense to sparse remains challenging.
Due to the reduction in the amount of activated parameters and the newly introduced gate network for expert routing, we observe a significant performance decline between the \llamoe models and the original dense LLaMA models.

To solve the above issues, we comprehensively explore four different methods for expert construction.
Among them, the non-overlapping randomly splitting method achieves the best performance.
Subsequently, we continue training the transformed MoE models and additional gate networks.
In this stage, we also carefully study both dynamic and static data sampling strategies for obtaining the fastest convergence and performance improvement.
Finally, with a static domain weight proportion corresponding to the activated parameters, the \llamoe models can quickly converge to a decent level with 200B tokens.

In summary, our contributions are as follows:
\begin{itemize}
\item We propose a framework to develop mixture-of-experts from existing decoder-style LLMs by splitting FFNs and continual pre-training, which has never been explored before.
\item  We comprehensively explore different splitting methods for expert construction. Meanwhile, we comprehensively investigate both dynamic and static data sampling strategy for continual pre-training.
\item  Our extensive experiments on a variety of tasks validate the effectiveness of our proposed \llamoe series models. Notably, all our model construction processes and training data are transparent.
\end{itemize}

\section{Related Work}

\textbf{Mixture-of-Experts (MoE).} 
Traditionally, dense models feed all parameters to each input token. In this way, the growing model capacity brings increased computational cost. To alleviate this issue, sparse models attempt to activate a subset of parameters for each input and these activated parameters are referred as experts.
In \cite{shazeer2017outrageously}, MoE was first proven effective in modern deep learning.
This work added an MoE layer which was stacked between LSTM, resulting in state-of-the-art results in language modeling and machine translation benchmarks.
Subsequently, the MoE layer is introduced to the transformer architecture as a substitute for the FFN layers.
GShard \citep{lepikhin2020gshard} applied the MoE to the Transformer and significantly improved machine translation across 100 languages.
Switch Transformers \citep{fedus2022switch} further scales the language model's size to the trillion-level parameter with a simple and effective MoE layer design. 
Naively trained MoE models is prone to load imbalance, e.g., only a few experts are frequently used while the others are scarcely activated. For optimizing the training, BASE layer \citep{lewis2021base}, HASH layer \citep{roller2021hash}, and Expert Choice \citep{zhou2022mixture} study how to build MoE models to fully utilize the model capacity.
Recently, for model architecture, \citet{xue2024openmoe} explore training a decoder-only MoE with a modified UL2 training objective.
Mixtral is another decoder-style MoE model that selects two out of eight experts with token-choice routing~\citep{ai_mixtral_2023}.

\textbf{Expert Construction.} There are two lines of works constructing MoE from dense checkpoints. The first category splits the parameters of the FFNs and ensures that the total model parameters remain unchanged.
MoEBERT \citep{zuo2022moebert} propose an
importance-based method to adapt the FFNs into experts. Considering that some neurons in the FFNs contribute more to the model performance, they share the most important neurons (i.e., the ones with the highest scores) among the experts, and the other neurons are distributed evenly. 
MoEfication \citep{zhang2021moefication} study the activation patterns of FFNs in Transformer models and find a sparse activation
phenomenon. Then, they discover the functional partitions (experts) in FFNs and build
routers for selecting experts. It is worth noting that they only focus on the ReLU-based FFNs in T5 \citep{raffel2020exploring} and BERT \citep{devlin2018bert}. 
There is another type of work that expands the total model parameters while keeping the activation parameters as the original dense models. 
Sparse upcycling \citep{komatsuzaki2022sparse} explore upgrading an existing dense model into a larger, sparsely activated MoE. In particular, the experts in the new MoE layer are identical copies of the original MLP layer that is replaced. In this paper, our work follows the first research line and decomposes the original FFNs into multiple small experts. Different from MoEBERT and MoEfication, our work focuses on a SwiGLU-based decoder-style models and continues training the MoE models.   

\section{Preliminary}

A standard Mixture of Experts (MoE) layer comprises $N$ expert networks $\left\{E_1, E_2, \dots, E_N\right\}$ and a gating network $G$ which activates the top-$k$ experts and distributes input tokens to corresponding experts. 
In general, the number of selected experts $k$ is fixed and much smaller than the total number of experts $N$, which presents the sparsely activated fashion of MoE models.
Formally, given an input embedding $x$, $E_i(x)$ denotes the output of the $i$-th expert network, the MoE layer's output is the sum of outputs from $k$ selected experts:
\begin{equation}
    y = \sum_{i \in \mathcal{K}} {G(x)_i \cdot E_i(x)},
\end{equation}
where the top-$k$ indices are determined by $G(x)$, indicating which experts accept the input $x$. $\mathcal{K}$ is the set of selected top-$k$ indices.
Following \cite{shazeer2017outrageously}, we implement a token-level noisy top-$k$ gating with load balancing in \llamoe. 

\section{Methodology}

\begin{figure*}[t]
    \centering
    \includegraphics[width=1.0\linewidth]{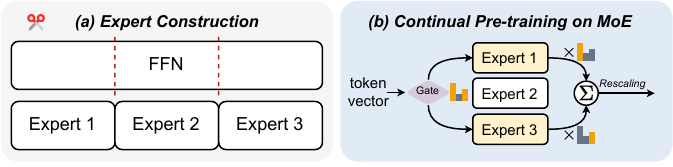}
    \caption{
    The main framework of building \llamoe models.
    (a) The original FFNs in the LLaMA are split into different experts. (b) In the transformed \llamoe, the hidden states are processed by partially chosen experts instead of all experts. We continue to train the \llamoe to improve the performance.}
    \label{fig:framework}
\end{figure*}

As illustrated in Figure~\ref{fig:framework}, we construct \llamoe from \llama-2-7B by first partitioning FFNs into multiple experts and routing each token to top-$k$ experts.
Continual pre-training is subsequently applied to recover the MoE model's language ability. In the following sections, 
we first introduce the expert construction method from the original dense model, then present the data sampling and processing strategies in continual pre-training.

\subsection{Expert Construction}\label{sec:expert-init}
In this section, we describe the construction of each expert network in the \llamoe. We start with the feed-forward network in LLaMA which uses SwiGLU \citep{shazeer2020glu} as the activation function. Each FFN layer in LLaMA consists of three parts: an up projection weight $W_\mathrm{up}\in\mathbb{R}^{d\times d_h}$, a gate projection weight $W_\mathrm{gate}\in\mathbb{R}^{d\times d_h}$ and a down projection weight $W_\mathrm{down}\in\mathbb{R}^{d_h\times d}$. Given an input $x\in\mathbb{R}^d$, the output $y\in\mathbb{R}^d$ of the FFN is:
\begin{equation}
        y = hW_\mathrm{down},\quad h = xW_\mathrm{up} \odot \mathrm{Swish}(xW_\mathrm{gate}).
\end{equation}

In \llamoe, each expert network is implemented as a feed-forward layer. 
Specifically, given the expert size $m$ and the selection indices set $S_j$, the weights of the $j$-th expert network $E_j$ is formulated as:
\begin{equation}
    \begin{aligned}
        W_\mathrm{up}^{(j)} = {W_\mathrm{up}}_{:,S_j} \in\mathbb{R}^{d\times m},\quad
        W_\mathrm{gate}^{(j)} = {W_\mathrm{gate}}_{:,S_j} \in\mathbb{R}^{d\times m},\quad
        W_\mathrm{down}^{(j)} = {W_\mathrm{down}}_{S_j,:} \in\mathbb{R}^{m\times d}, \label{eq:weight}
    \end{aligned}
\end{equation}
where the selection indices set $S_j$ is:
\begin{equation}
    S_j = \left\{i_{j_1}, i_{j_2}, \dots, i_{j_m} \ | \ 1 \leq i_{j_1} \neq i_{j_2} \neq \cdots \neq i_{j_m} \leq d_h \right\}.
\end{equation}

Given an input $x\in\mathbb{R}^d$, the output $E_j(x)\in\mathbb{R}^d$ of the $j$-th expert network $E_j$ is:
\begin{equation}
        E_j(x) = h_jW_\mathrm{down}^{(j)},\quad h_j = xW_\mathrm{up}^{(j)} \odot \mathrm{Swish}(xW_\mathrm{gate}^{(j)}).
\end{equation}

Based on whether the intermediate neurons within the FFN are shared among different experts, we implement two groups of construction methods: \textit{Neuron-Independent} and \textit{Neuron-Sharing}. 

\textbf{Neuron-Independent}. We formulate expert construction as a task of partitioning into equal-sized sets. Given a universal set $U$ containing indices of all intermediate neurons $\left\{1,2,\dots,d_h\right\}$, we uniformly split $U$ into $n$ equal-sized indices set $S_1,S_2,\dots,S_n$ and construct experts with size $m=\frac{d_h}{n}$ according to Equation \ref{eq:weight}, where we have:
\begin{equation}
    \begin{aligned}
        \bigcup_{i=1}^{n}S_i = U \quad\text{and}\quad \bigcap_{i=1}^{n}S_i = \varnothing. \\
    \end{aligned}
\end{equation}
Specifically, we describe two kinds of partition methods: 
\begin{itemize}
    \item \textbf{Independent}$_\text{Random}$: We randomly partition $U$ into $n$ equal-sized subsets.
    \item \textbf{Independent}$_\text{Clustering}$: Following \cite{zhang2021moefication}, we perform a balanced k-means clustering \citep{balanced_kmeans} with $n$ centroids on the row vectors of $W_\mathrm{up}$ and partition $U$ according to the clustering result.
\end{itemize}
    
\textbf{Neuron-Sharing}. According to \cite{zuo2022moebert}, the representation ability of a model can be partially retained through a structured partition.
Therefore, we treat the expert construction as a structured pruning problem, by measuring the first-order Taylor expansion on loss change $\Delta L$ for each intermediate neuron when it gets pruned. For each FFN layer, we maintain a vector $v\in\mathbb{R}^{d_h}$ initialized as zeros to record the importance of its intermediate neurons. Given batched data $D$, the importance vector $v$ is updated as follows:
\begin{equation}
    v := v + \sum_{(x,y)\in D} \big\lvert h \odot \nabla_{h} L(x,y) \big\rvert. \label{text:importance}
\end{equation}
The indices sets $S_1,S_2,\dots,S_n$ are then generated using certain algorithm for the experts with sizes $m=\frac{d_h}{n}$. Given the universal indices set $U=\left\{1,2,\dots,d_h\right\}$, we have:
\begin{equation}
    \bigcup_{i=1}^{n}S_i \in U.
\end{equation}

\begin{itemize}
    \item \textbf{Sharing}$_\text{Inner}$: We obtain $n$ importance vectors $v_1,v_2,\dots,v_n$ through pre-clustered $n$ groups of data. For each expert $i$, the corresponding $S_i$ consists the indices of neurons with the largest $m$ values in $v_i$.
    \item \textbf{Sharing}$_\text{Inter}$: Referencing the implementation in \cite{rajbhandari2022deepspeed}, we set aside the neurons shared by most experts as independent residual blocks, while others are assigned according to the importance vectors $v_1,v_2,\dots,v_n$.
\end{itemize}

\paragraph{Re-scaling}
After partitioning a dense FFN layer into multiple small experts, the activated expert parameters are much smaller than the original dense models. To preserve the representational capacity of the partitioned model, we introduce a scale factor and apply rescale operations to guarantee effective expert output. In particular, considering activating $k$ out of $N$ experts, we will scale the output of expert by a factor of $\frac{N}{k}$.

\begin{table}[]
    \renewcommand{\arraystretch}{1.1}
    \centering
    \begin{tabular}{lcccc}
        \toprule
        Model & \#Activated Experts & \#Total Experts & \#Activated Params \\
        \midrule
        OPT-2.7B & - & - & 2.7B \\
        Pythia-2.8B & - & - & 2.8B \\
        INCITE-Base-3B & - & - & 2.8B \\
        Open-LLaMA-3B-v2 & - & - & 3.4B \\
        Sheared-LLaMA-2.7B & - & - & 2.7B \\
        \midrule
        \llamoe-3.0B (2/16) & 2 & 16 & 3.0B \\
        \llamoe-3.5B (4/16) & 4 & 16 & 3.5B \\
        \llamoe-3.5B (2/8) & 2 & 8 & 3.5B \\
        \bottomrule
    \end{tabular}
    \vspace{8pt}
    \caption{
        The statistics for model parameters and activation parameters for sparse MoE models. All \llamoe-3.0B and \llamoe-3.5B models have the same parameters as LLaMA-2-7B. \llamoe-3.5B has two variants including 2/8 and 4/16. They have different numbers of experts but the same amount of activation parameters. 
    }
    \label{tab:model_structure}
\end{table}

\subsection{Continual Pre-training}

Since the original \llama model structure is reorganized after converting to MoE, we continue pre-training the \llamoe model in Table~\ref{tab:model_structure} to recover their language modeling abilities.
The training objective is the same as \llama-2~\citep{touvron2023llama}.
To improve the training efficiency, we explore different data sampling strategies and data quality filtering methods as follows.

\paragraph{Data Sampling Weights.}
The data sampling weights are crucial to obtain a global optimum~\citep{xie2023doremi}.
LLaMA-v1 utilizes a set of static empirical sampling weights~\citep{touvron2023llama}, while some of the domains (e.g. Wikipedia) have been proven to be less effective on downstream tasks~\citep{shen2023slimpajamadc}.
This indicates it may be not appropriate to assign large weights when sampling these domains.
\cite{xie2023doremi} employ additional models to obtain better static sampling weights.
Although it is faster to get convergence, it brings additional training compute.
\cite{xia2023sheared} introduce a dynamic weight sampling strategy in the training phase, which boosts performance on downstream tasks.

To obtain better performances, we investigate the following data sampling strategies for \llamoe continual pre-training.
Data sampling weights are adjusted every 2.5B tokens in dynamic settings and the total training budget is 30B tokens.

\begin{itemize}
    \item \textbf{Static}$_{\text{\llama}}$: Training with static \llama-1 sampling weights.
    \item \textbf{Static}$_{\text{Sheared}}$: Applying final static sampling weights of Sheared-LLaMA.
    \item \textbf{Dynamic}$_{\text{\llama}}$: Sheared-LLaMA dynamic sampling with \llama-v1 weights construction. We evaluate LLaMA-v2 on a subset of SlimPajama with all the training domains for obtaining the reference loss.
    \item \textbf{Dynamic}$_{\text{Uniform}}$: Sheared-LLaMA dynamic sampling with uniform weights construction.
\end{itemize}

\paragraph{Data Filtering.}
As our training budget is limited, we further explore two data filter strategies to speed up model convergence.
Specifically, we filter out $\sim$50\% advertisements and $\sim$15\% non-fluent texts in CommonCrawl and C4 datasets.

\section{Experiments}

\subsection{Training Dataset}
The training dataset for \llamoe is SlimPajama \citep{cerebras2023slimpajama}, which cleans and deduplicates the RedPajama dataset.
This dataset contains 627B tokens and encompasses data from seven domains, including CommonCrawl, C4, Github, Wikipedia, Books, arXiv, and StackExchange. 

\subsection{Evaluation Datasets and Comparing Models}

According to \cite{wei2023skywork} and \cite{deepseek-llm}, the performance on HellaSwag~\citep{zellers2019hellaswag} grows smoothly during pre-training.
A similar trend is also found in ARC-c~\citep{allenai_arc}, thus we utilize HellaSwag and ARC-c as the evaluation datasets for the analysis experiments.

For comprehensive model ability assessment, we follow \cite{xia2023sheared} and use the lm-evaluation-harness \citep{eval-harness} to evaluate the following downstream tasks: 0-shot {normalized accuracy} (acc\_norm) of ARC easy \citep{clark2018think}, LAMBADA \citep{paperno2016lambada}, LogiQA \citep{liu2020logiqa}, PIQA \citep{bisk2020piqa}, SciQ \citep{welbl2017crowdsourcing}, and WinoGrande Standard \citep{sakaguchi2021winogrande}, 10-shot HellaSwag \citep{zellers2019hellaswag}, 25-shot ARC Challenge \citep{clark2018think}, and 5-shot MMLU \citep{hendrycks2020measuring}. If there is no normalized accuracy, we use acc instead. Furthermore, we use OpenCompass~\citep{2023opencompass} to evaluate 32-shot NQ~\citep{2019nq}.
We compare \llamoe with strong pre-trained language models containing similar activation parameters, including OpenLLaMA-3B-v2~\citep{openlm2023openllama}, OPT-2.7B~\citep{opt}, Pythia-2.8B~\citep{biderman2023pythia}, INCITE-Base-3B~\citep{together2023redpajama}, and Sheared-LLaMA~\citep{xia2023sheared}.

\begin{table}[t]

\renewcommand{\arraystretch}{1.1} 
\setlength{\tabcolsep}{3pt}
\resizebox{\textwidth}{!}{
\begin{tabular}{lcccccc}
    \toprule
     & \multicolumn{6}{c}{\textbf{Commonsense \& Reading Comprehension}} \\ 
    \cmidrule(lr){2-7} 
    \textbf{Model} & \textbf{SciQ} & \textbf{PIQA} & \textbf{WinoGrande} & \textbf{ARC-E} & \textbf{ARC-C (25)} & \textbf{HellaSwag (10)} \\
    
    \midrule
        OPT-2.7B & 78.9 & 74.8 & 60.8 & 54.4 & 34.0 & 61.4 \\
        Pythia-2.8B & 83.2 & 73.6 & 59.6 & 58.8 & 36.7 & 60.7 \\
        INCITE-Base-3B & 85.6 & 73.9 & 63.5 & 61.7 & 40.3 & 64.7 \\
        Open-LLaMA-3B-v2 & 88.0 & \textbf{77.9} & 63.1 & 63.3 & 40.1 & 71.4 \\
        Sheared-LLaMA-2.7B & 87.5 & 76.9 & 65.0 & 63.3 & 41.6 & 71.0 \\ 
        \hdashline
        \llamoe-3.0B & 84.2 & 77.5 & 63.6 & 60.2 & 40.9 & 70.8 \\ 
        \llamoe-3.5B (4/16) & 87.6 & \textbf{77.9} & 65.5 & \textbf{65.6} & \textbf{44.2} & \textbf{73.3} \\ 
        \llamoe-3.5B (2/8) & \textbf{88.4} & 77.6 & \textbf{66.7} & 65.3 & 43.1 & \textbf{73.3} \\ 
        
    \midrule 
    
     & \multicolumn{2}{c}{\textbf{Continued}} & \textbf{LM} & \multicolumn{2}{c}{\textbf{World Knowledge}} & \\
    \cmidrule(lr){2-3} \cmidrule(lr){4-4} \cmidrule(lr){5-6} 
    \textbf{Model} & \textbf{LogiQA} & \textbf{BoolQ (32)} & \textbf{LAMBADA} & \textbf{NQ (32)} & \textbf{MMLU (5)} & \multirow{-2}{*}{\textbf{Average}} \\
    \midrule
        OPT-2.7B & 25.8 & 63.3 & 63.6 & 10.7 & 25.8 & 50.3 \\
        Pythia-2.8B & 28.1 & 65.9 & 64.6 & 8.7 & 26.8 & 51.5 \\
        INCITE-Base-3B & 27.5 & 65.8 & 65.4 & 15.2 & 27.2 & 53.7 \\
        Open-LLaMA-3B-v2 & 28.1 & 69.2 & 67.4 & 16.0 & 26.8 & 55.6 \\
        Sheared-LLaMA-2.7B & 28.3 & 73.6 & 68.3 & 17.6 & \textbf{27.3} & 56.4 \\ 
        \hdashline
        \llamoe-3.0B & \textbf{30.6} & 71.9 & 66.6 & 17.0 & 26.8 & 55.5 \\
        \llamoe-3.5B (4/16) & 29.7 & \textbf{75.0} & \textbf{69.5} & \textbf{20.3} & 26.8 & \textbf{57.7} \\
        \llamoe-3.5B (2/8) & 29.6 & 73.9 & 69.4 & 19.8 & 27.0 & 57.6 \\
    
    \bottomrule
\end{tabular}
}

\vspace{8pt}

\caption{
Main results on downstream tasks.
We re-evaluate all the models on these datasets.
LLaMA-MoE-3.5B significantly outperforms publicly available models of comparable size on most downstream tasks.
The shot number used is noted in parentheses, with 0-shot if not specified.
}

\label{tab:main_result}
\end{table}

\begin{figure*}[t]
    \centering
    \begin{subfigure}{0.32\textwidth}
        \centering
        \includegraphics[width=\linewidth]{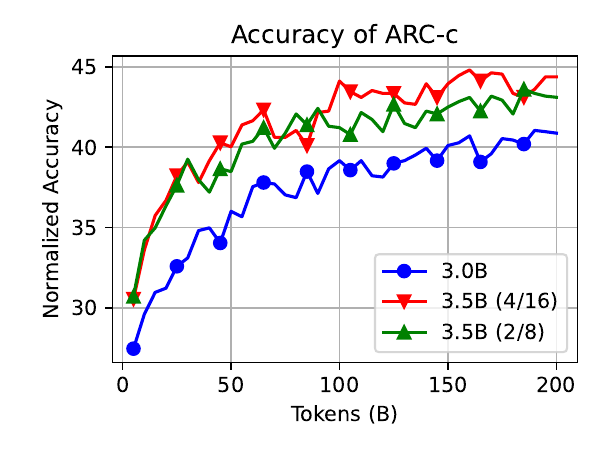}
        \caption{ARC-c (25)}
        \label{fig:main-results-arc}
    \end{subfigure}
    \begin{subfigure}{0.32\textwidth}
        \centering
        \includegraphics[width=\linewidth]{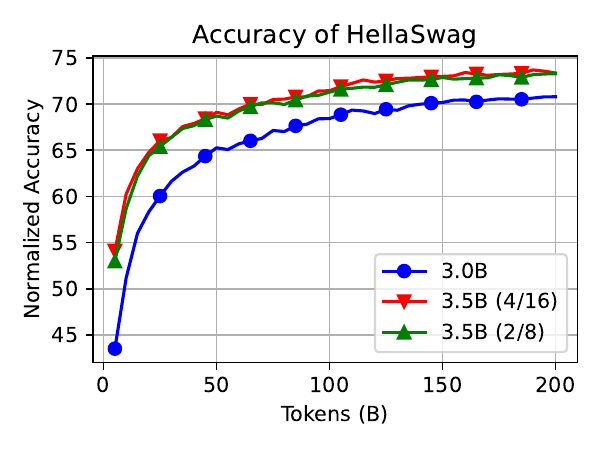}
        \caption{HellaSwag (10)}
        \label{fig:main-results-hellaswag}
    \end{subfigure}
    \begin{subfigure}{0.32\textwidth}
        \centering
        \includegraphics[width=\linewidth]{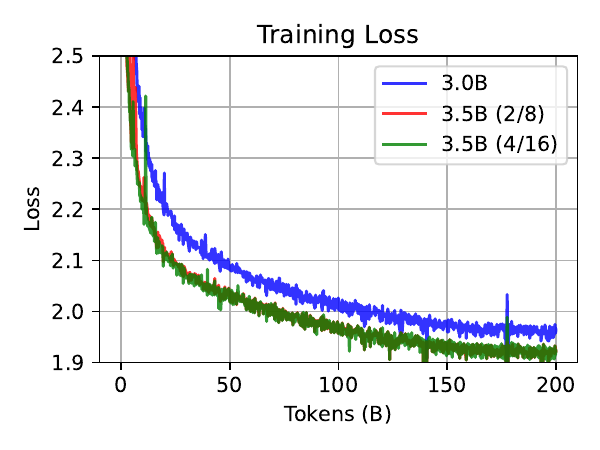}
        \caption{Training loss}
        \label{fig:main-results-loss}
    \end{subfigure}
    \caption{Model performances on ARC-c and HellaSwag dataset and the training loss for \llamoe-3.0B and \llamoe-3.5B. The two models are trained with 200B tokens.}
    \label{fig:main-results}
\end{figure*}

\subsection{Experiment Settings}

We start from \llama-2-7B~\citep{llama2} and explore different \moe construction strategies.
All models are trained on 112 A100 (80G) GPUs with a global batch size of 15M tokens. The context length is 4096.
The maximum learning rate is 2e-4 with 100 warmup steps and the final learning rate decays to 2e-5 with cosine scheduling.
Each \llamoe variant is expected to be trained on 200B tokens (13.6k steps).
Our implementation is based on transformers~\citep{transformers}, ZeRO-1~\citep{rajbhandari2022deepspeed}, and FlashAttention v2~\citep{flashattention2}. 
The final \llamoe models are trained on \textbf{Independent}$_\text{Random}$ with \textbf{Static}$_\text{Sheared}$ data sampling weights and fluency-filtered SlimPajama datasets.
More details can be found in our released code.

\subsection{Main Results}

As shown in Table~\ref{tab:main_result}, \llamoe-3.5B (2/8) and \llamoe-3.5B (4/16) achieve similar average results and the latter is slightly better. However, 
\llamoe-3.5B significantly surpasses open-source models with similar activation parameters. Specifically, \llamoe-3.5B (4/16) exceeds the most competitive model Sheared-LLaMA by 1.3 average points. 
Meanwhile, \llamoe-3.0B performs comparably with Open-LLaMA-3B-v2. To demonstrate the training progress and model capability changes.
In Figure~\ref{fig:main-results-arc} and \ref{fig:main-results-hellaswag}, we present the model performances on both ARC-c and HellaSwag and find the results on these datasets grow gradually as the training process goes on.
There are more fluctuations in ARC-c results, while HellaSwag provides smoother results. For the training loss, as shown in Figure~\ref{fig:main-results-loss}, \llamoe-3.0B and \llamoe-3.5B converges to about 1.95 and 1.90, respectively. The final loss are higher than LLaMA-2 7B as these two models activate relatively fewer parameters. 

\subsection{Expert Construction}

\begin{figure}
    \centering
    \begin{subfigure}{0.32\textwidth}
        \centering
        \includegraphics[width=\linewidth]{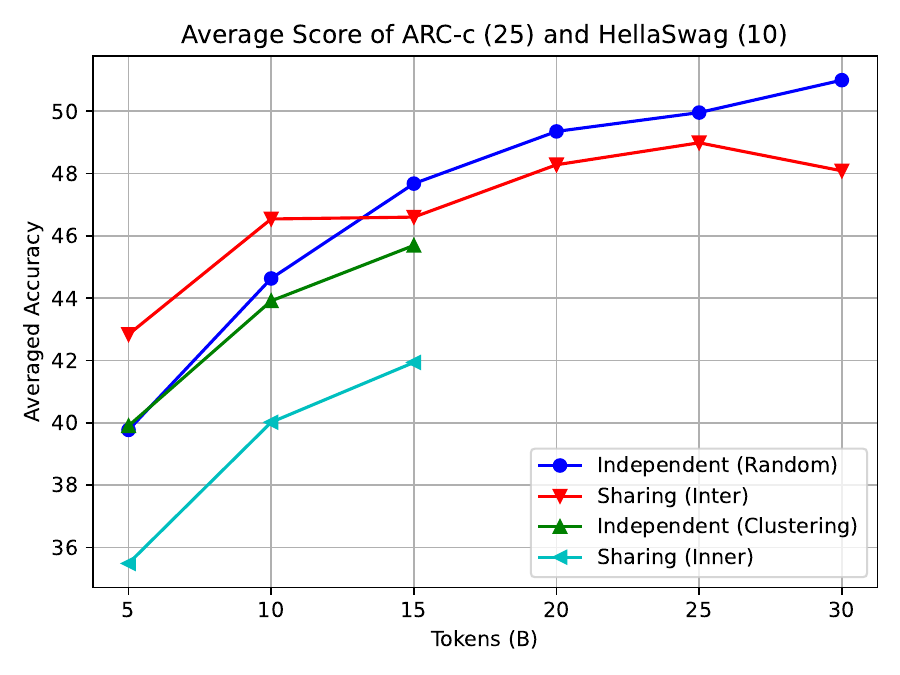}
        \caption{Average score}
        \label{fig:expert-init-score}
    \end{subfigure}
    \begin{subfigure}{0.32\textwidth}
        \centering
        \includegraphics[width=\linewidth]{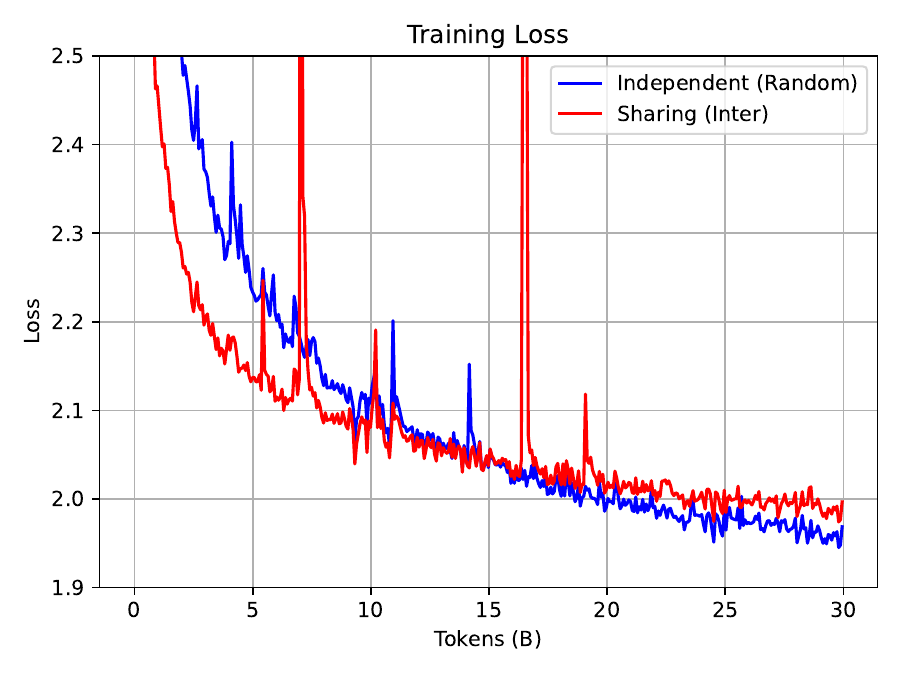}
        \caption{Training loss}
        \label{fig:expert-init-loss}
    \end{subfigure}
    \begin{subfigure}{0.33\textwidth}
        \centering
        \includegraphics[width=\linewidth]{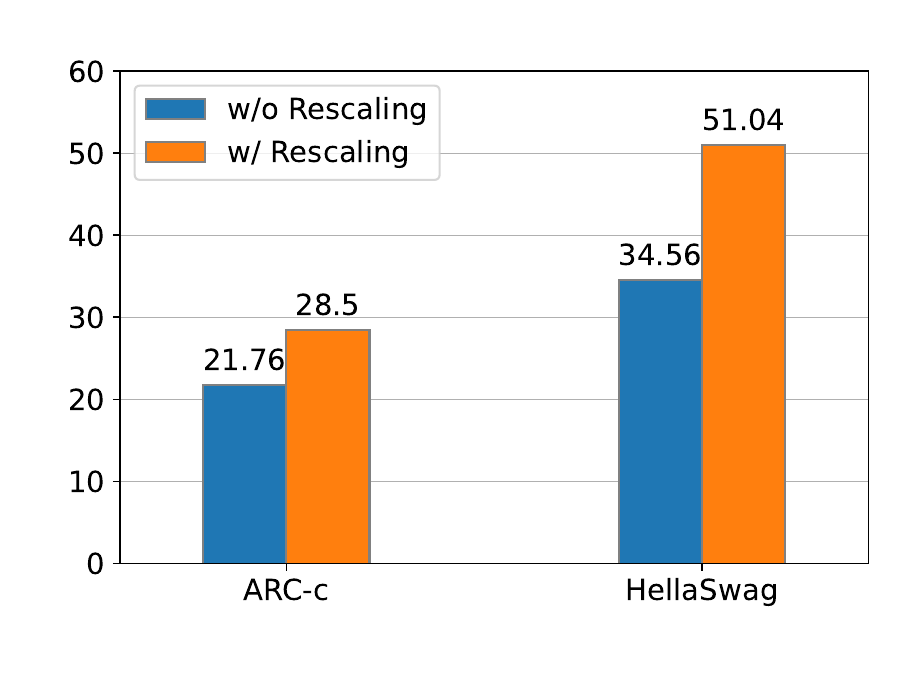}
        \caption{Re-scaling}
        \label{fig:ablation-scale-factor}
    \end{subfigure}
    \caption{Model performances with different expert construction methods. Among four kinds of construction methods, \textbf{Independent}$_\text{Random}$ obtains the best result.
    We also present the ablation study of expert output re-scaling after 5B tokens of continual pre-training.
    }
    \label{fig:expert-init-results}
\end{figure}

In this section, we compare four types of expert construction methods as introduced in \S~\ref{sec:expert-init}.
Interestingly, as presented in Figure~\ref{fig:expert-init-results}, \textbf{Independent}$_\text{Random}$ achieves the best average score within the token budget.
Since gates and experts are trained simultaneously, other partition methods may bring bias when construction, which introduces additional difficulties for recovering the model's language ability by continual pre-training.
\textbf{Sharing}$_\text{Inter}$ has a good convergence trend at the first 5B tokens, but it struggles to get better loss performance.
Actually, we have trained more tokens for \textbf{Sharing}$_\text{Inter}$ and the results are significantly lower than \textbf{Independent}$_\text{Random}$.
We can also observe changes in the loss value as depicted in Figure~\ref{fig:expert-init-loss}.
However, We found that models should be trained for at least 15$\sim$20B tokens to properly conclude. 
For \textbf{Sharing}$_\text{Inner}$, an average of $\sim50\%$ neurons are shared between each expert pair, thus the upper bound for this variant is quite low and the model would like to achieve lower performance.
Finally, as \textbf{Independent}$_\text{Clustering}$ and \textbf{Sharing}$_\text{Inner}$ achieves much lower performances than other methods, so we only train those models for 15B tokens.

Besides the specific expert construction strategies, we conduct the ablation study on the re-scaling operation on the expert outputs.
As shown in Figure~\ref{fig:ablation-scale-factor}, we find that the re-scaling on the expert outputs provides significantly better performance for MoE models.

\subsection{Data Sampling Weights}

\begin{figure}
    \centering
    \begin{subfigure}{0.49\textwidth}
        \centering
        \includegraphics[width=\linewidth]{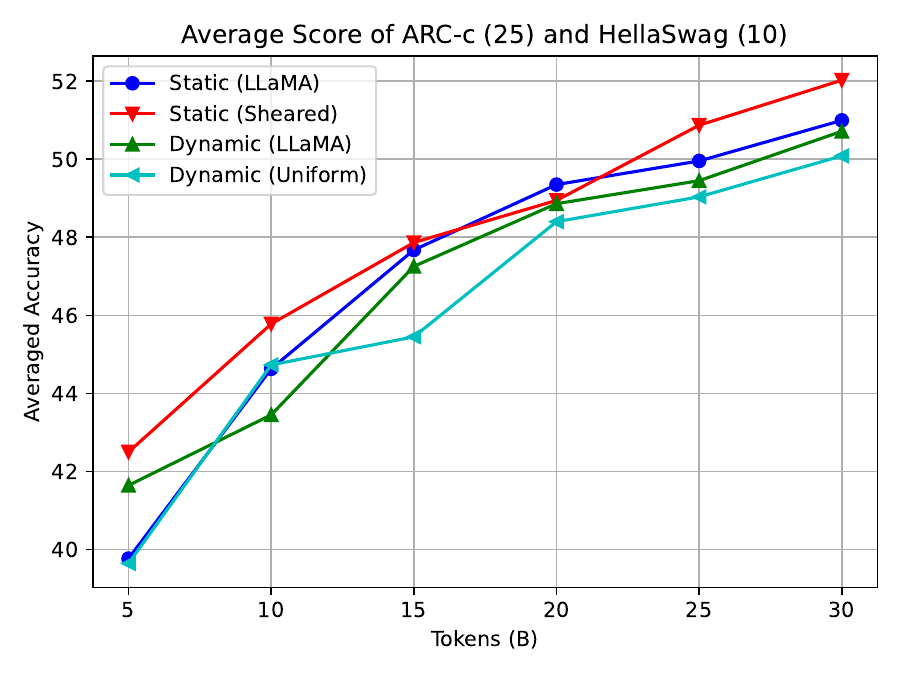}
        \caption{Average score}
        \label{fig:data-sampling-weights-score}
    \end{subfigure}
    \begin{subfigure}{0.49\textwidth}
        \centering
        \includegraphics[width=\linewidth]{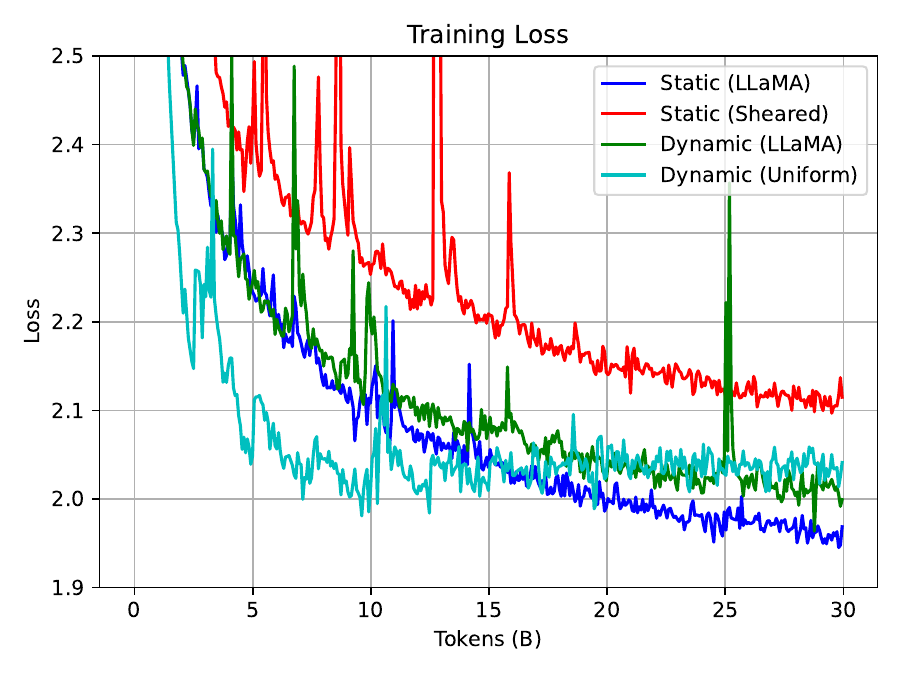}
        \caption{Training loss}
        \label{fig:data-sampling-weights-loss}
    \end{subfigure}
    \caption{Model performances with different data sampling strategies. Among four sampling ways, Static$_{\text{Sheared}}$ achieves the best performance. However, it does not achieve the lowest training loss.}
    \label{fig:data-sampling-weights}
\end{figure}

\begin{figure}
    \centering
    \begin{subfigure}{0.24\textwidth}
        \centering
        \includegraphics[width=\linewidth]{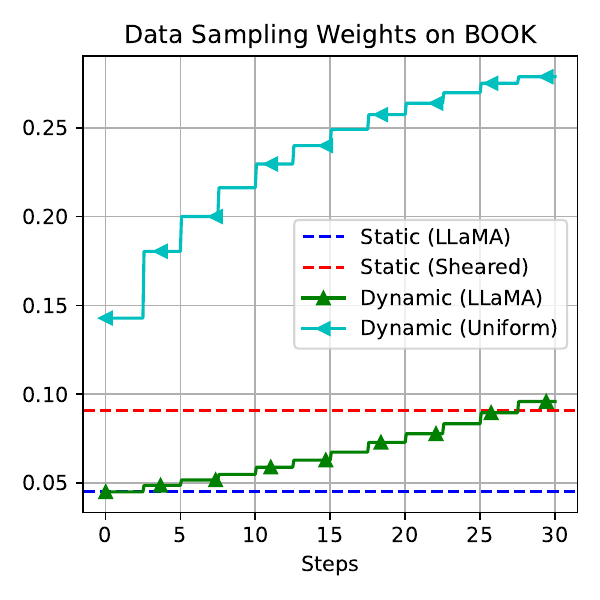}
        \caption{Book}
        \label{fig:data-sampling-weights-book}
    \end{subfigure}
    \begin{subfigure}{0.24\textwidth}
        \centering
        \includegraphics[width=\linewidth]{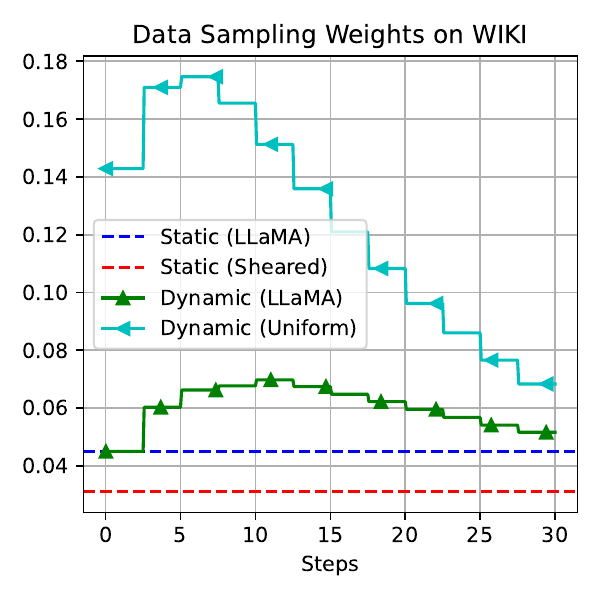}
        \caption{Wikipedia}
        \label{fig:data-sampling-weights-wiki}
    \end{subfigure}
    \begin{subfigure}{0.24\textwidth}
        \centering
        \includegraphics[width=\linewidth]{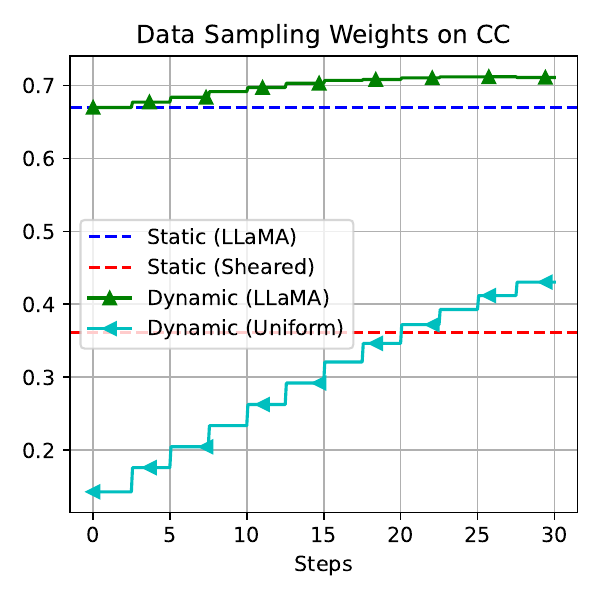}
        \caption{CommonCrawl}
        \label{fig:data-sampling-weights-cc}
    \end{subfigure}
    \begin{subfigure}{0.24\textwidth}
        \centering
        \includegraphics[width=\linewidth]{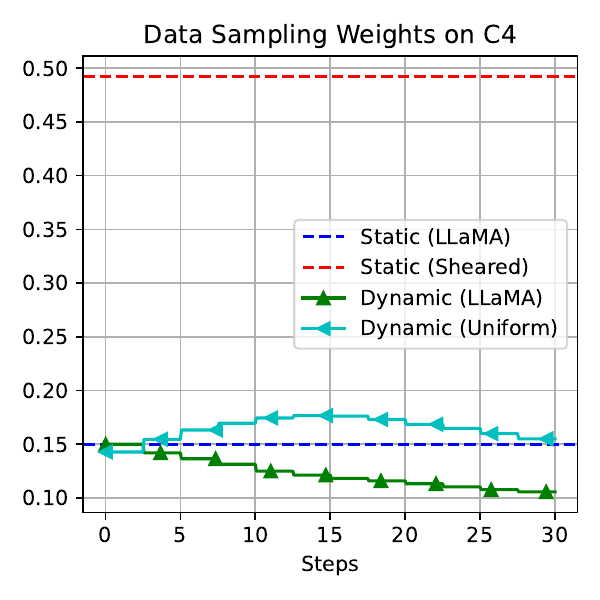}
        \caption{C4}
        \label{fig:data-sampling-weights-c4}
    \end{subfigure}
    \caption{Data sampling weights variation on four domains. For Static$_{\text{Sheared}}$ and Static$_{\text{LLaMA}}$, the sampling weight is fixed among the training process, while the domain importance gradually changes for Dynamic$_{\text{Uniform}}$ and Dynamic$_{\text{LLaMA}}$. Both Dynamic$_{\text{Uniform}}$ and Dynamic$_{\text{LLaMA}}$ are two dynamic weight sampling strategies from \citep{xia2023sheared}.}
    \label{fig:data-sampling-weights-variation}
\end{figure}

As Figure~\ref{fig:data-sampling-weights-score} shows, Static$_{\text{Sheared}}$ surpasses other methods within the token budget, and dynamic data sampling weights are worse than static weights.
However, the Static$_{\text{Sheared}}$ loss in Figure~\ref{fig:data-sampling-weights-loss} is greater than other methods, which indicates that the continual pre-training loss may be less relevant to downstream task performances.
The loss of Dynamic$_{\text{Uniform}}$ drops down quickly, but it suffers from the instability problem and contains many fluctuations.
From Figure~\ref{fig:data-sampling-weights-variation}, we find the sampling weight of C4 goes to the opposite directions compared to Static$_{\text{Sheared}}$ because the estimated Sheared LLaMA-2.7B reference loss is lower than LLaMA2-7B (2.033 vs. 2.075).
It is very tricky to select the best reference loss, we leave it for future work.

\subsection{Data Filtering}

\begin{figure}
    \centering
    \begin{subfigure}{0.49\textwidth}
        \centering
        \includegraphics[width=\linewidth]{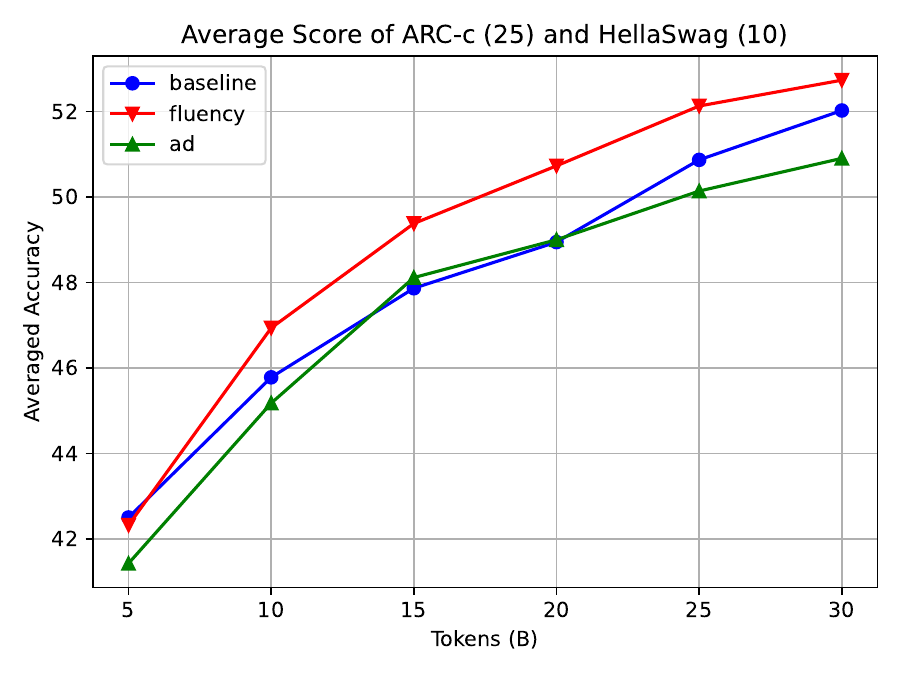}
        \caption{Average score}
        \label{fig:data-filtering-score}
    \end{subfigure}
    \begin{subfigure}{0.49\textwidth}
        \centering
        \includegraphics[width=\linewidth]{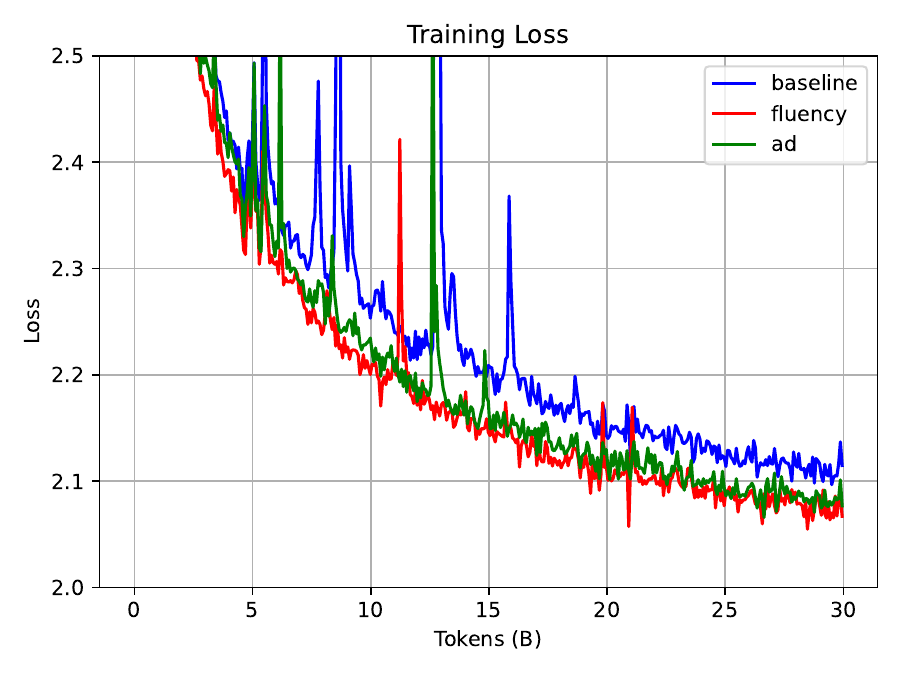}
        \caption{Training loss}
        \label{fig:data-filtering-loss}
    \end{subfigure}
    \caption{
        Model performances with different data filtering strategies.
        Here, the ``baseline" is \textbf{Independent}$_\text{Random}$ together with \textbf{Static}$_\text{Sheared}$ data sampling weights. ``fluency" and ``ad" means the baseline strategy is equipped with removing non-fluent texts or ads. 
    }
    \label{fig:data-filtering}
\end{figure}

We further filter out data with advertisements and those in low fluency.
The results are presented in Figure~\ref{fig:data-filtering}.
Both fluency and advertisement filtering obtain lower training loss than the baseline.
However, advertisement filtering perform worse in downstream tasks.
We think the number of filtered advertisements are too large to bring more knowledge and information, and the filtering tagger should be improved with fine-grained thresholding adjustment.
The fluency filtering method successfully removes texts in low-quality and improves the average score.
Based on the results, we train our final model with the fluency-filtered dataset.
It is worth noting that we do not introduce any new datasets but remove part of them considering model convergence speed.

\subsection{\llamoe vs. Training from Scratch}

\begin{figure}[t]
    \centering
    \begin{subfigure}{0.49\textwidth}
        \centering
        \includegraphics[width=\linewidth]{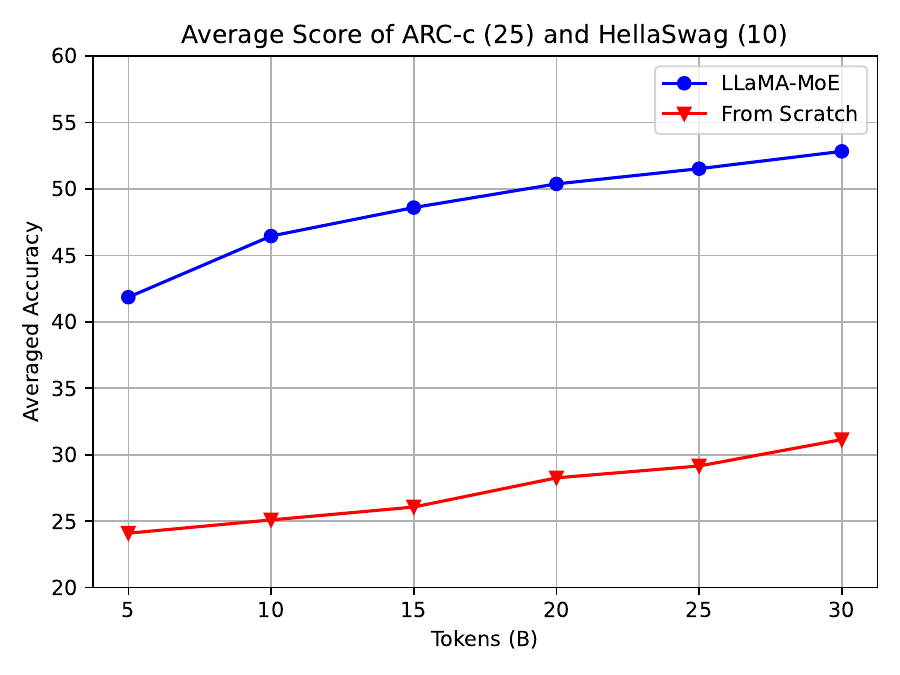}
        \caption{Average score}
        \label{fig:scratch-average-score}
    \end{subfigure}
    \begin{subfigure}{0.49\textwidth}
        \centering
        \includegraphics[width=\linewidth]{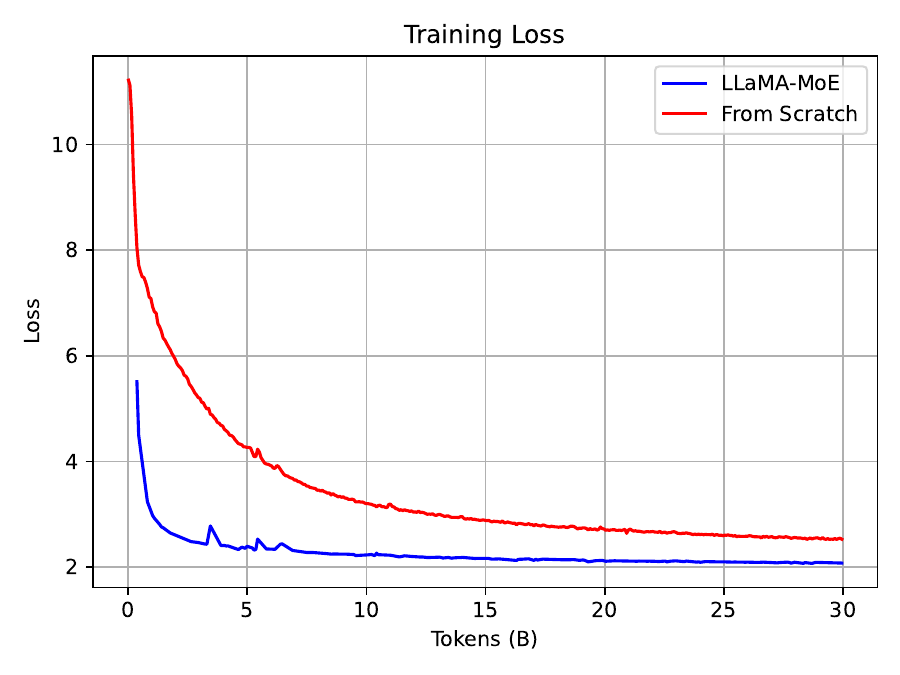}
        \caption{Training loss}
        \label{fig:scratch-loss}
    \end{subfigure}
    \caption{
    Model performances on \llamoe-3.5B (2/8) and MoE training from scratch.
    For comparing the initial losses, we set the y-axis limit to a larger scale.
    }
    \label{fig:from-scratch}
\end{figure}

To validate the effectiveness of our proposed LLaMA-MoE, we conduct experiments on training the MoE model from scratch.
By comparing the training losses in Figure~\ref{fig:scratch-loss} and performances in Figure~\ref{fig:scratch-average-score}, we find that constructing MoE models from dense models has significant advantages and could reduce huge computing resources.

\subsection{Expert Specialization}

\begin{figure}
    \centering
    \begin{subfigure}{0.24\textwidth}
        \centering
        \includegraphics[width=\linewidth]{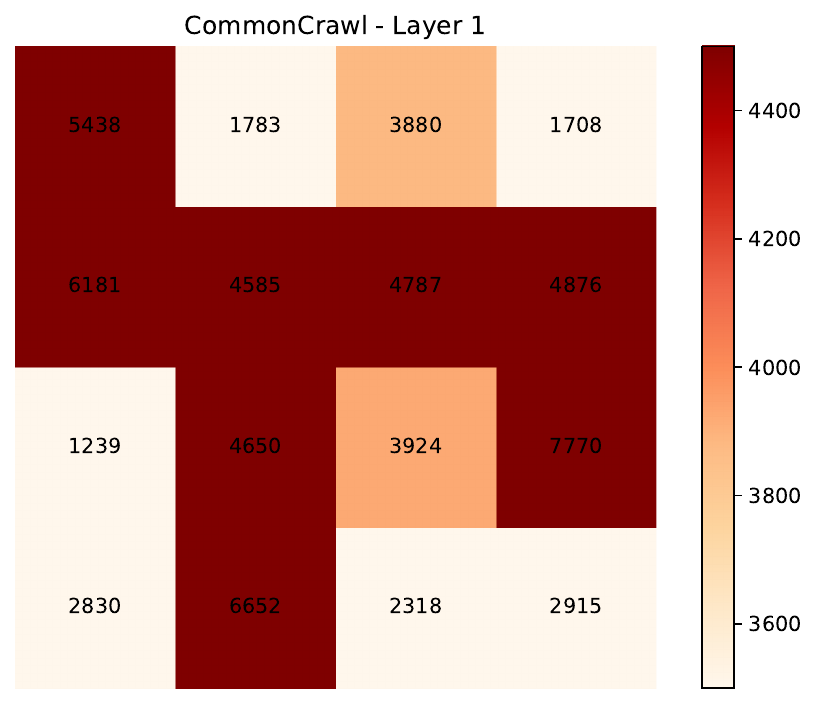}
        \caption{CommonCrawl (1)}
        \label{fig:expert-specialization-cc1}
    \end{subfigure}
    \begin{subfigure}{0.24\textwidth}
        \centering
        \includegraphics[width=\linewidth]{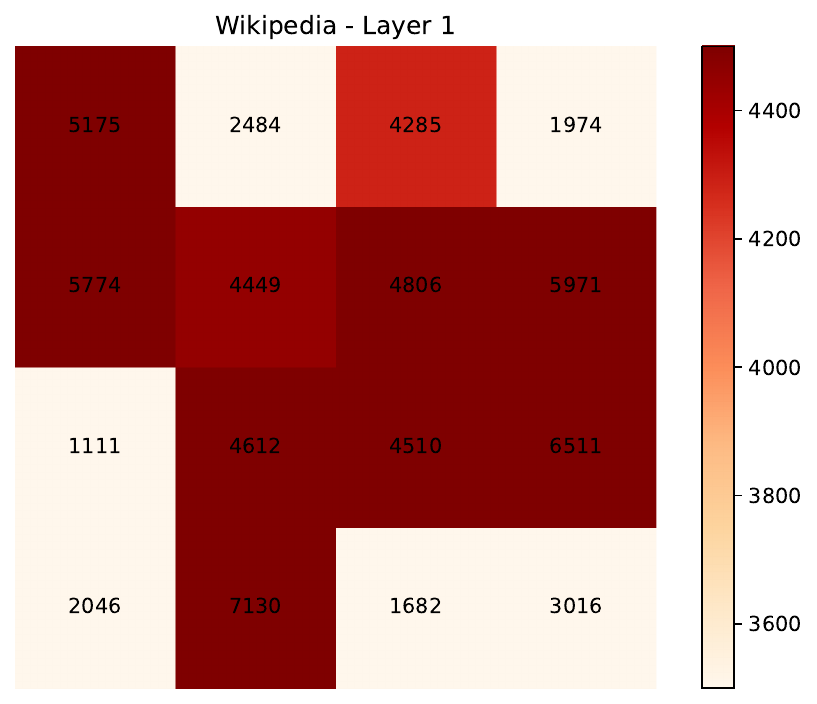}
        \caption{Wikipedia (1)}
        \label{fig:expert-specialization-wiki1}
    \end{subfigure}
    \begin{subfigure}{0.24\textwidth}
        \centering
        \includegraphics[width=\linewidth]{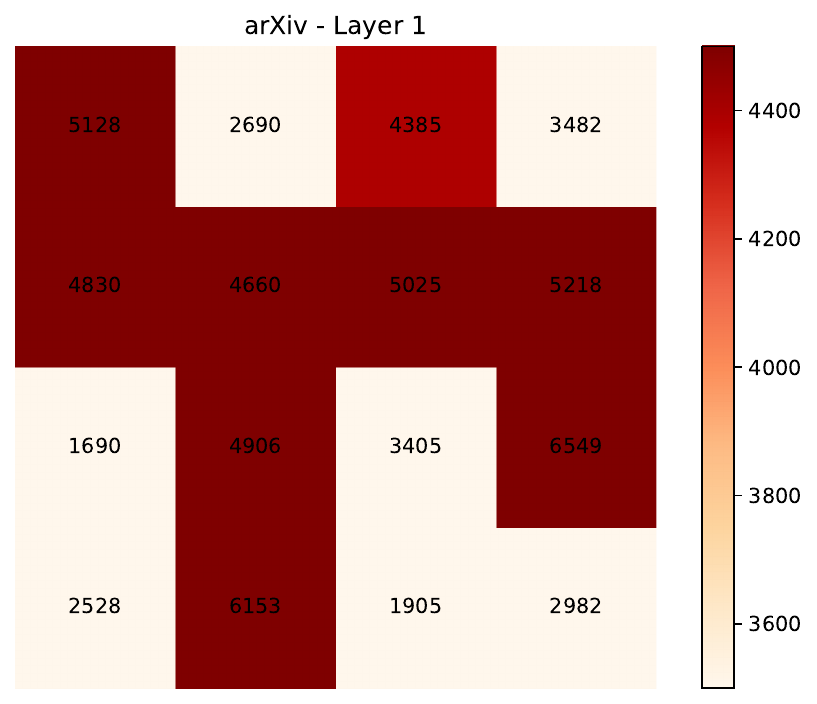}
        \caption{arXiv (1)}
        \label{fig:expert-specialization-arxiv1}
    \end{subfigure}
    \begin{subfigure}{0.24\textwidth}
        \centering
        \includegraphics[width=\linewidth]{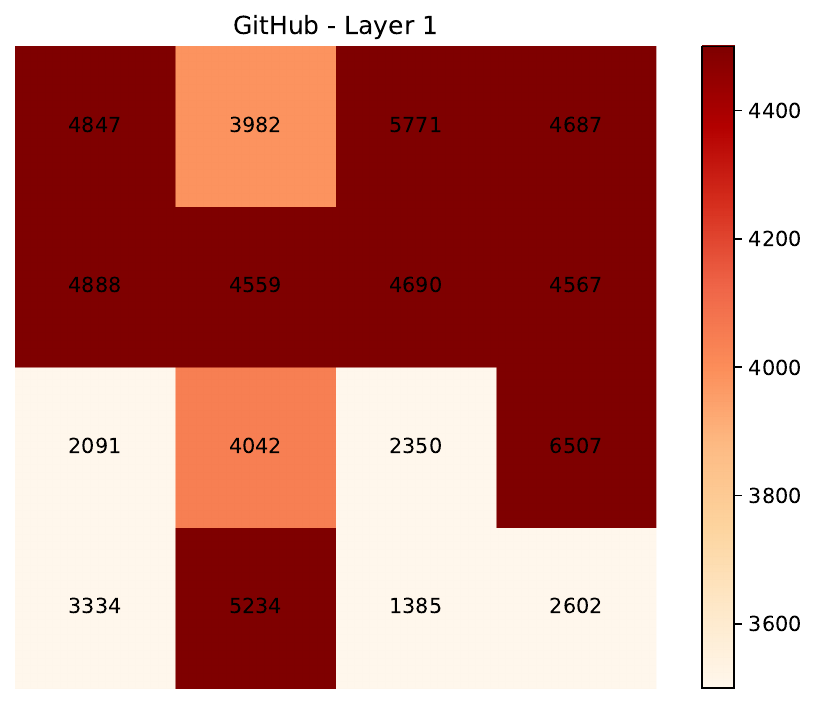}
        \caption{GitHub (1)}
        \label{fig:expert-specialization-github1}
    \end{subfigure}

    \begin{subfigure}{0.24\textwidth}
        \centering
        \includegraphics[width=\linewidth]{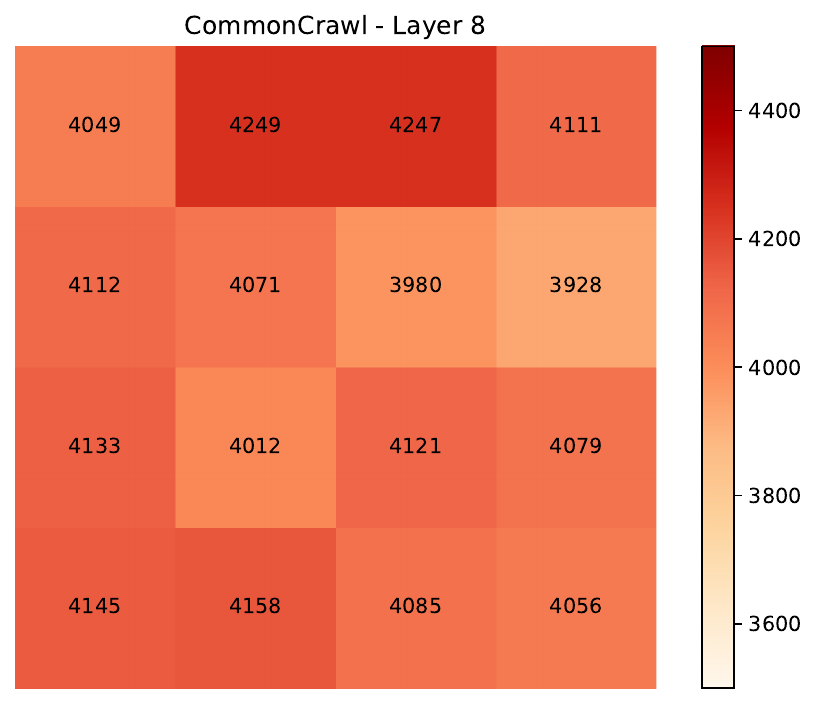}
        \caption{CommonCrawl (8)}
        \label{fig:expert-specialization-cc8}
    \end{subfigure}
    \begin{subfigure}{0.24\textwidth}
        \centering
        \includegraphics[width=\linewidth]{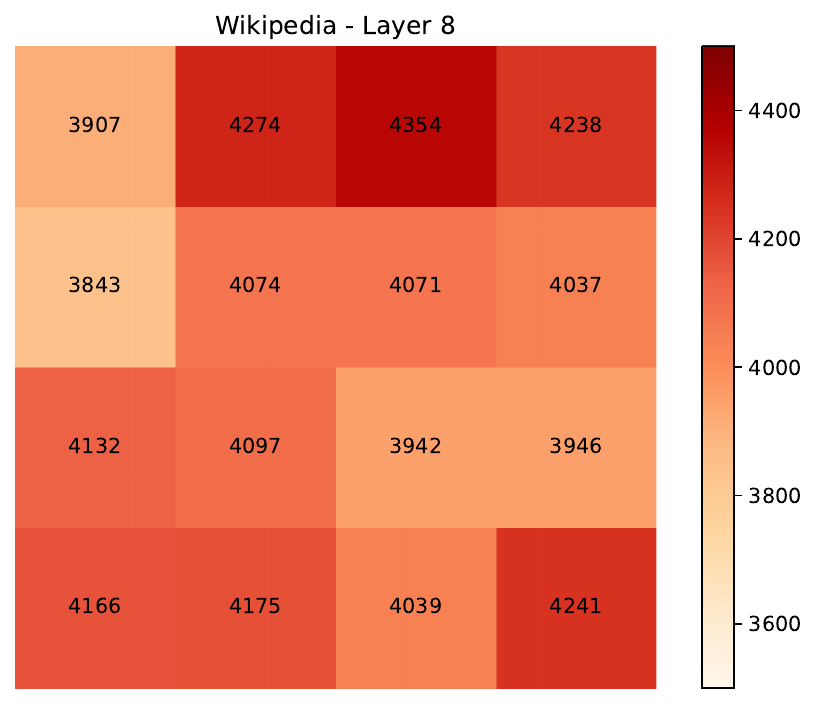}
        \caption{Wikipedia (8)}
        \label{fig:expert-specialization-wiki8}
    \end{subfigure}
    \begin{subfigure}{0.24\textwidth}
        \centering
        \includegraphics[width=\linewidth]{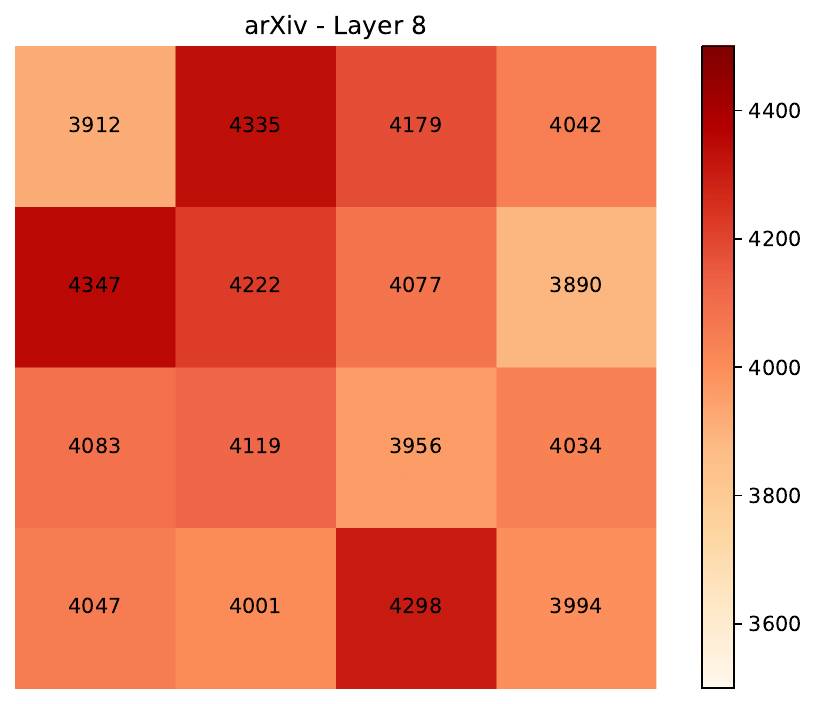}
        \caption{arXiv (8)}
        \label{fig:expert-specialization-arxiv8}
    \end{subfigure}
    \begin{subfigure}{0.24\textwidth}
        \centering
        \includegraphics[width=\linewidth]{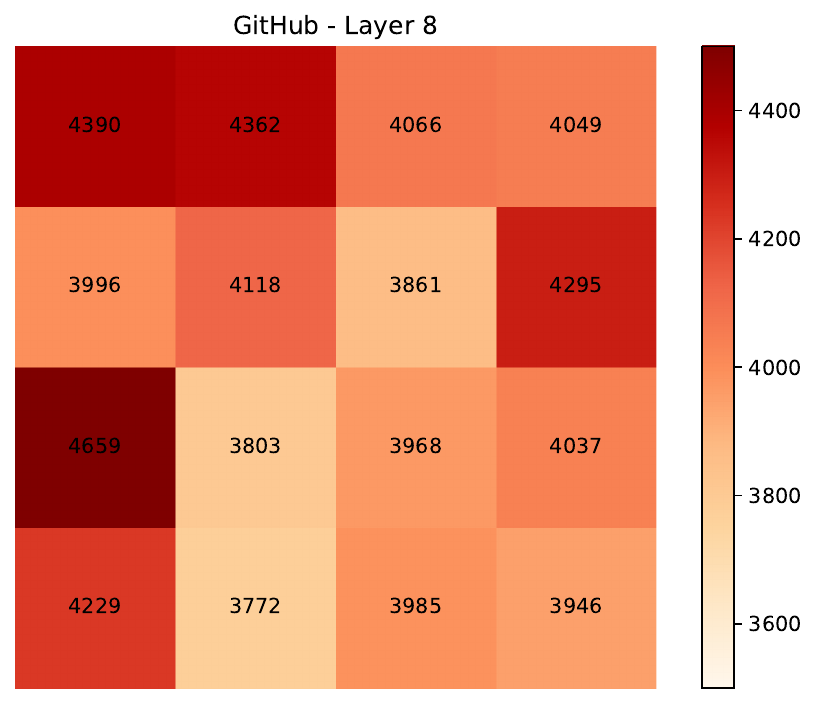}
        \caption{GitHub (8)}
        \label{fig:expert-specialization-github8}
    \end{subfigure}

    \begin{subfigure}{0.24\textwidth}
        \centering
        \includegraphics[width=\linewidth]{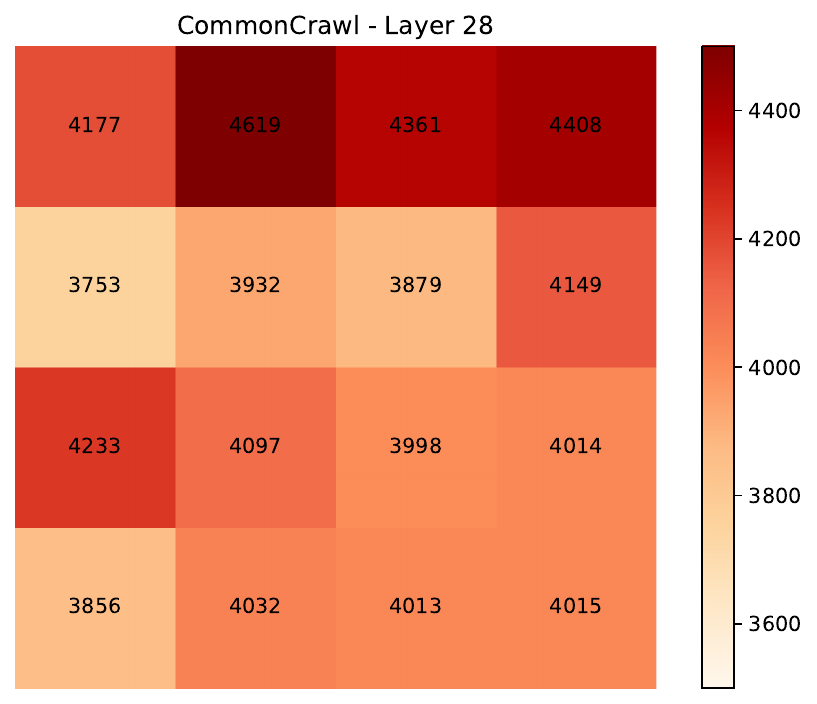}
        \caption{CommonCrawl (28)}
        \label{fig:expert-specialization-cc28}
    \end{subfigure}
    \begin{subfigure}{0.24\textwidth}
        \centering
        \includegraphics[width=\linewidth]{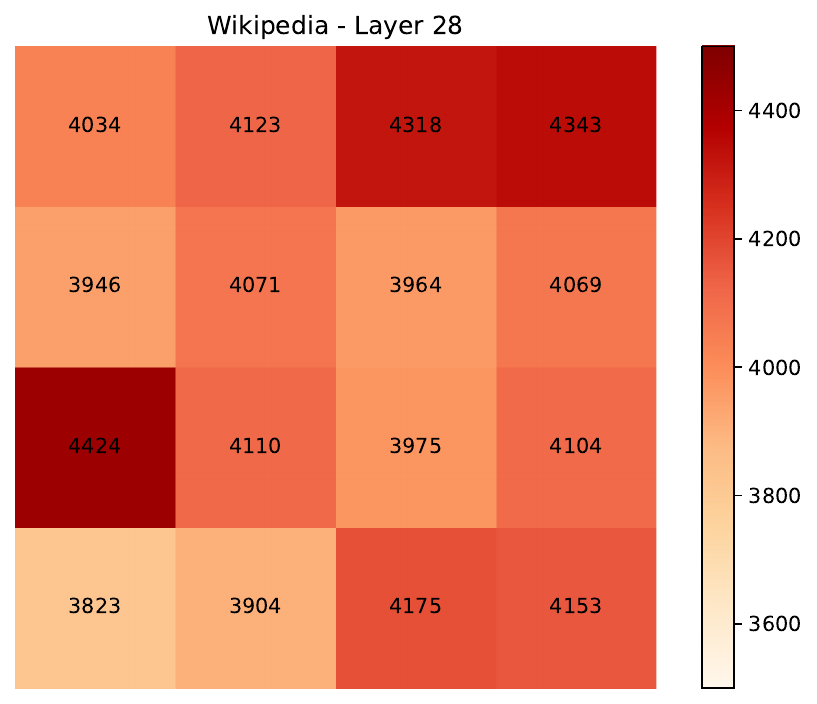}
        \caption{Wikipedia (28)}
        \label{fig:expert-specialization-wiki28}
    \end{subfigure}
    \begin{subfigure}{0.24\textwidth}
        \centering
        \includegraphics[width=\linewidth]{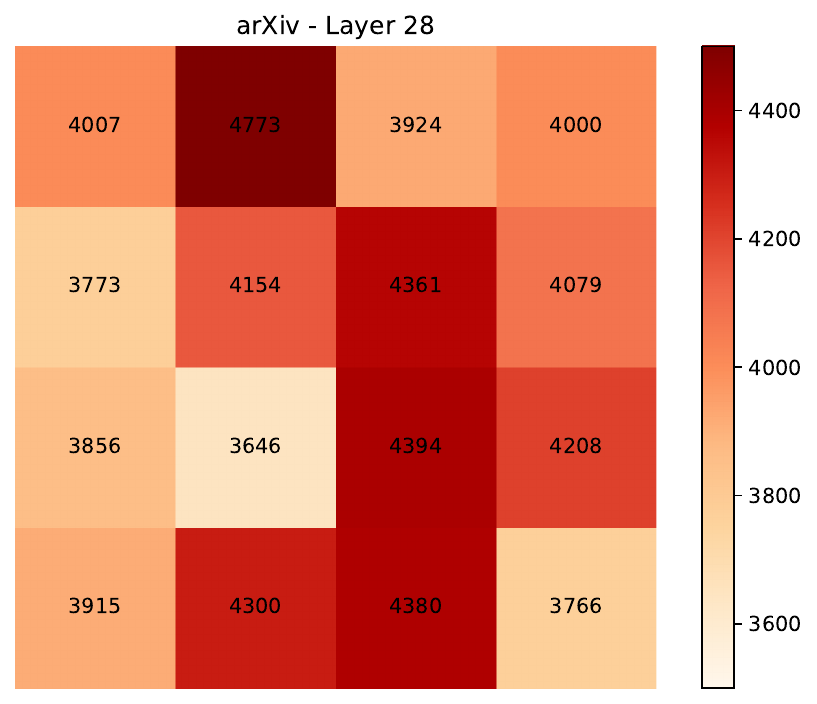}
        \caption{arXiv (28)}
        \label{fig:expert-specialization-arxiv28}
    \end{subfigure}
    \begin{subfigure}{0.24\textwidth}
        \centering
        \includegraphics[width=\linewidth]{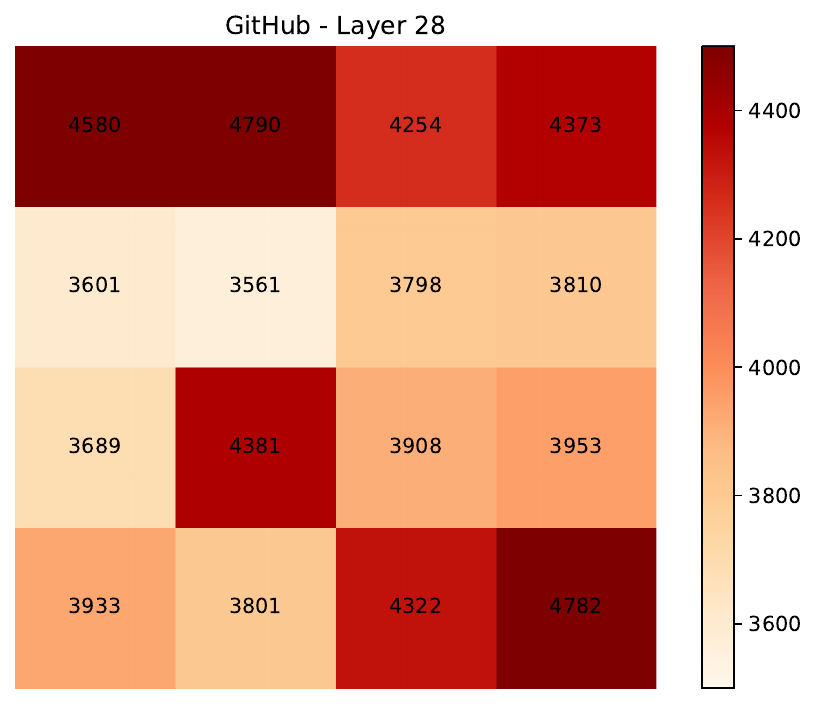}
        \caption{GitHub (28)}
        \label{fig:expert-specialization-github28}
    \end{subfigure}

    \begin{subfigure}{0.24\textwidth}
        \centering
        \includegraphics[width=\linewidth]{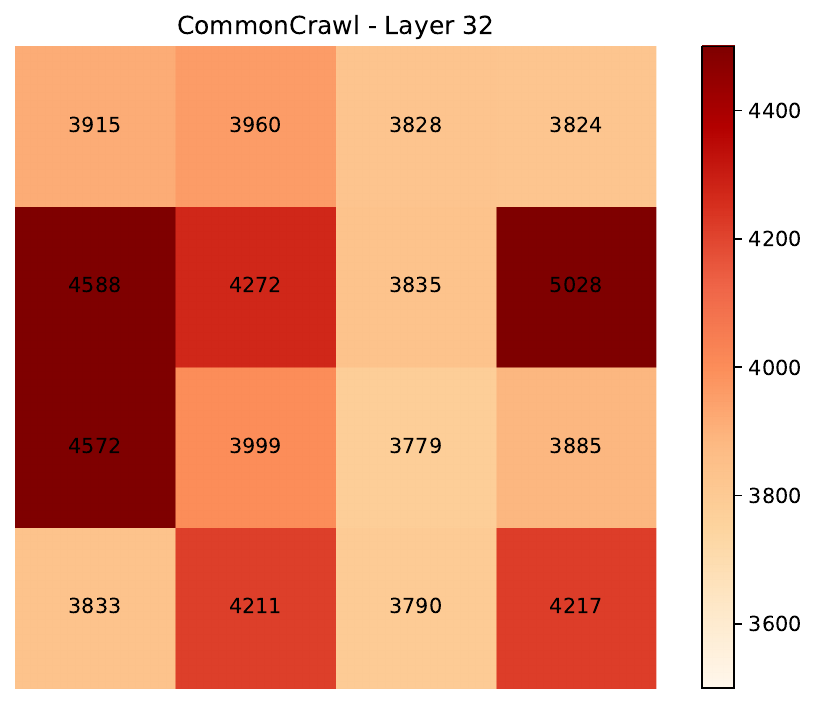}
        \caption{CommonCrawl (32)}
        \label{fig:expert-specialization-cc32}
    \end{subfigure}
    \begin{subfigure}{0.24\textwidth}
        \centering
        \includegraphics[width=\linewidth]{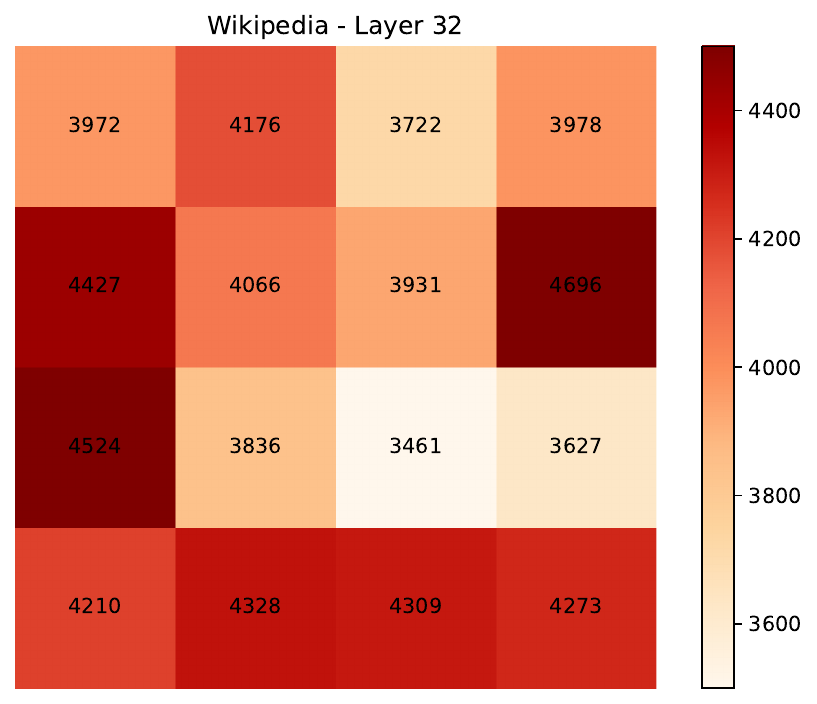}
        \caption{Wikipedia (32)}
        \label{fig:expert-specialization-wiki32}
    \end{subfigure}
    \begin{subfigure}{0.24\textwidth}
        \centering
        \includegraphics[width=\linewidth]{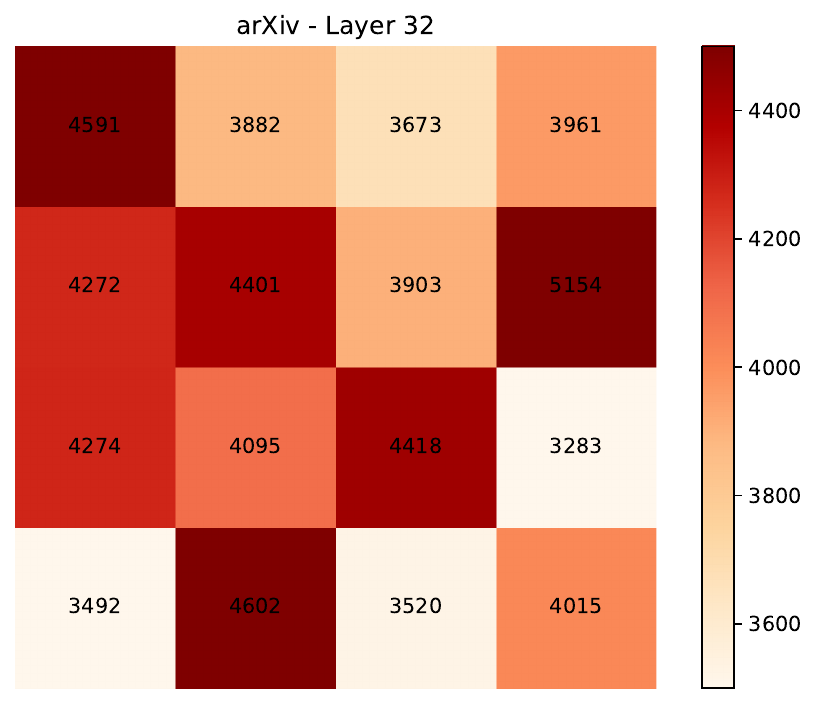}
        \caption{arXiv (32)}
        \label{fig:expert-specialization-arxiv32}
    \end{subfigure}
    \begin{subfigure}{0.24\textwidth}
        \centering
        \includegraphics[width=\linewidth]{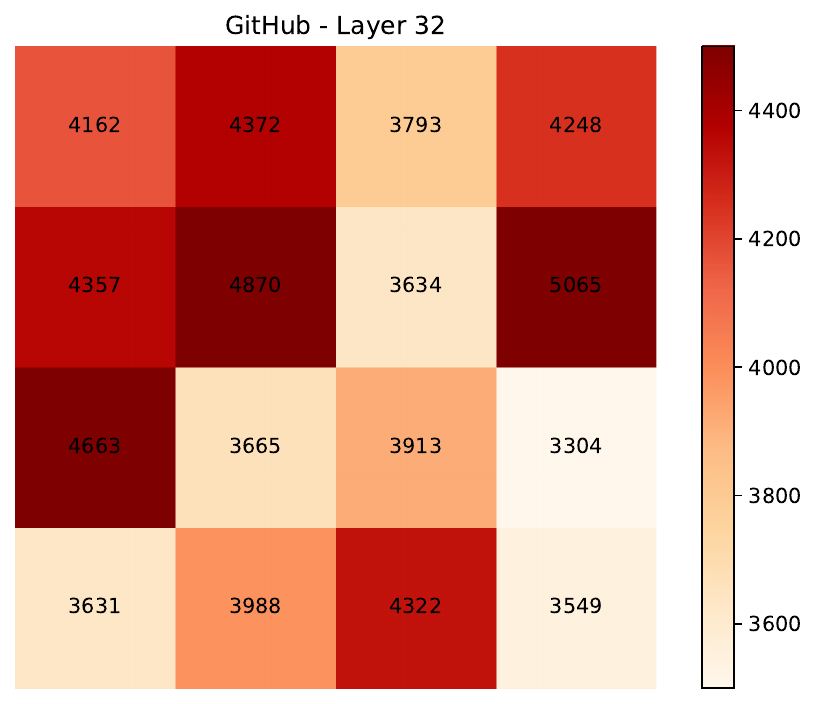}
        \caption{GitHub (32)}
        \label{fig:expert-specialization-github32}
    \end{subfigure}

    \caption{Expert routing statistics on the 1st, 8th, 28th, and 32nd layers for \llamoe-3.5B (4/16). Each cell represents the number of routed tokens to an expert. Our model has a total of 16 experts. 
    We sample 65.5K tokens from each domain for this visualization.}
    \label{fig:expert-specialization}
\end{figure}

As Figure~\ref{fig:expert-specialization} shows, deep layers have more routing preferences than shallow layers.
This may indicate that the shallow layers may capture more common features, while deep layers focus more on task-specific features.
Based on this finding, expert partition on the latter layers' FFNs may bring further improvements. We leave it for future exploration.
In deeper layers, each expert has different domain preferences and some experts are shared across different domains.
These shared experts may represent data similarities among different domains.
We also find the imbalance problem at the first two layers, where some experts are seldom selected.
These experts may be pruned for future MoE model compression.

\begin{figure}
    \centering
    \begin{subfigure}{0.49\textwidth}
        \centering
        \includegraphics[width=\linewidth]{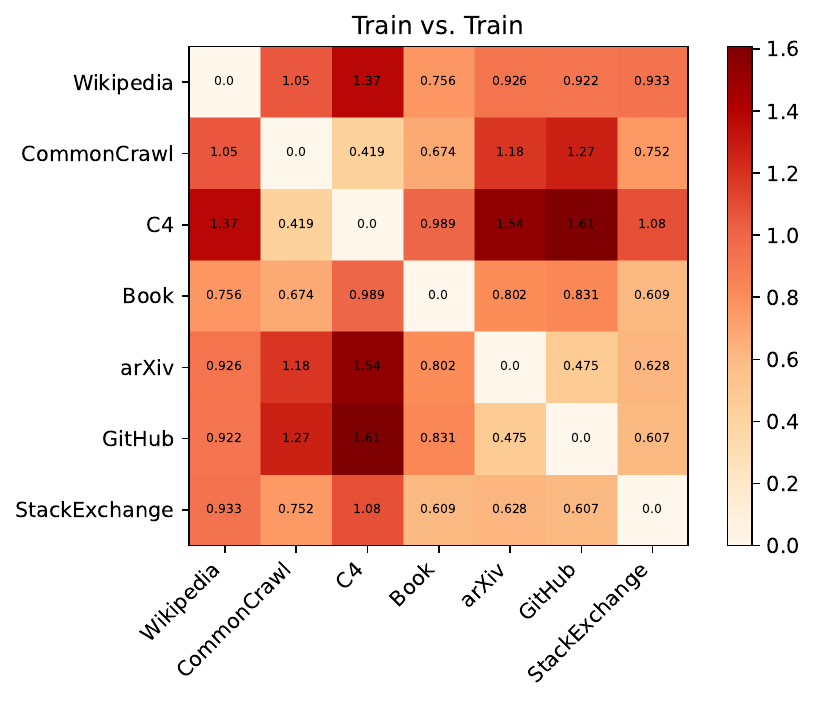}
        \caption{Train-Train}
        \label{fig:expert-routing-diff-train-train}
    \end{subfigure}
    \begin{subfigure}{0.49\textwidth}
        \centering
        \includegraphics[width=\linewidth]{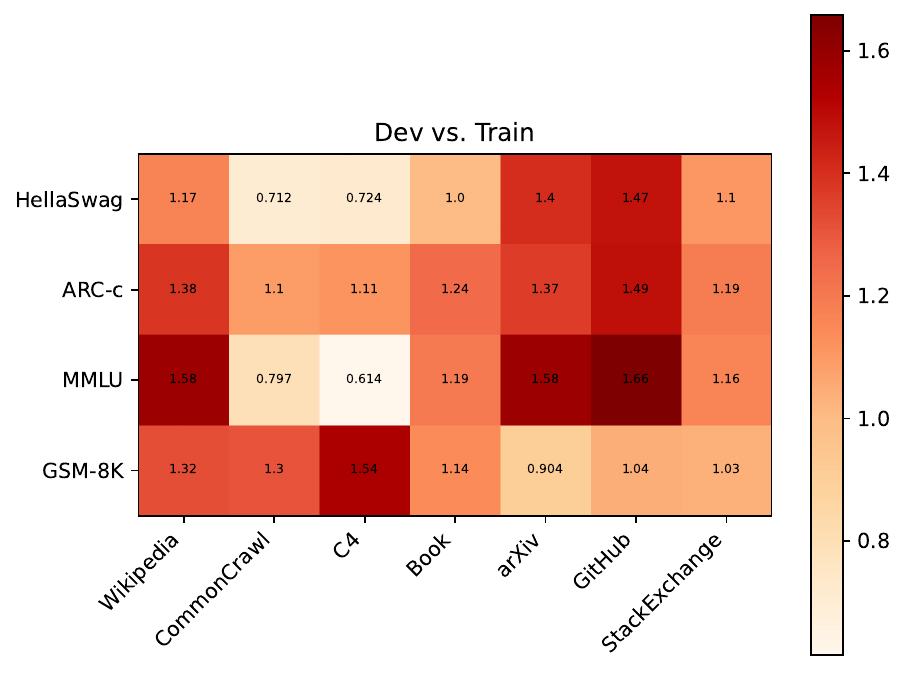}
        \caption{Dev-Train}
        \label{fig:expert-routing-diff-dev-train}
    \end{subfigure}
    \caption{
        Expert routing differences at the 32nd layer.
        Smaller numbers and lighter colors represent more similar expert routing preferences.
        8.4M tokens per domain are sampled for this experiment.
    }
    \label{fig:expert-routing-diff}
\end{figure}

To investigate the latent correlations among domains, we normalize the number of routed tokens and calculate the L2 distances to represent the expert selection differences.
As illustrated in Figure~\ref{fig:expert-routing-diff-train-train}, CommonCrwal and C4 datasets have similar expert preferences, while GitHub has similar expert preferences with arXiv and StackExchange.
As to the Dev-to-Train differences in Figure~\ref{fig:expert-routing-diff-dev-train}, we find HellaSwag and ARC-c share the most similar expert preferences with CommonCrawl and C4, and GSM-8K is similar to arXiv.
This may provide some insights for continual pre-training to further improve downstream performances.
For example, the model may consume more tokens from arXiv to improve GSM-8K results.
However, expert selections on ARC-c and GSM-8K have greater distances with current pre-training data, which may involve new domains to deal with such tasks.

\subsection{Instruction Tuning}

\begin{table*}[t]
\resizebox{\textwidth}{!}{%
\centering
\begin{tabular}{lcccccc}
\toprule
& \multicolumn{5}{c}{\textbf{Open LLM Leaderboard}} & \multicolumn{1}{c}{\textbf{Alignment}} \\
\cmidrule(lr){2-6} \cmidrule(lr){7-7}
\multirow{-2}{*}{\textbf{Model}} &
  \textbf{MMLU} &
  \textbf{ARC-C} &
  \textbf{HellaSwag} &
  \textbf{TruthfulQA} &
  \textbf{Average} &
  \textbf{MT-Bench} \\
\midrule

Sheared-LLaMA-2.7B-ShareGPT & \textbf{28.41} & 41.04 & 71.21 & 47.65 & 47.08 & 3.79 \\
Sheared-LLaMA-2.7B (Our Dataset) & 25.24 & 43.69 & 71.70 & \textbf{49.00} & 47.41 & 4.06 \\
LLaMA-MoE-v1-3.0B (2/16)    & 23.61 & 43.43 & 72.28 & 44.24 & 45.89 & 4.15 \\
LLaMA-MoE-v1-3.5B (4/16)    & 26.49 & \textbf{48.29} & \textbf{75.10} & 45.91 & \textbf{48.95} & 4.60 \\
LLaMA-MoE-v1-3.5B (2/8)     & 25.53 & 45.99 & 74.95 & 44.39 & 47.71 & \textbf{4.72} \\

\bottomrule
\end{tabular}
}%
\caption{Supervised fine-tuned model performances on Open LLM Leaderboard tasks and open-ended questions.
Sheared LLaMA-2.7B-ShareGPT is a chat model created by \citet{xia2023sheared}.
We reimplement the chat model by instruction tuning on our dataset and provide fair comparisons.
}
\label{tab:sft-results}
\end{table*}

MoE models have shown great prowess in instruction tuning with excellent task scaling properties~\citep{Shen2023MixtureofExpertsMI,chen2024textttmoerbench,zhu-et-al-2024-dynamic-sft-for-moe}.
To evaluate the instructed MoE models' performances, we fine-tune \llamoe with 6k curated ShareGPT instruction data~\citep{Liu2023DeitaWhatMG} for 2 epochs.
As shown in Table~\ref{tab:sft-results}, the instructed \llamoe-3.5B (4/16) outperforms the dense model on ARC-C (48.29 vs. 43.69) and HellaSwag (75.10 vs. 71.70) tasks.
The overall performance on Open LLM Leaderboard~\footnote{\url{https://huggingface.co/spaces/HuggingFaceH4/open_llm_leaderboard}} tasks surpasses the dense model (48.95 vs. 47.41).
Besides, there is a large gap in alignment abilities, where \llamoe-3.5B (2/8) outperforms Sheared LLaMA-2.7B by 0.66 scores on MT-Bench.

\section{Conclusion}
In this technical report, we build \llamoe-3.0B and \llamoe-3.5B models based on pre-trained \llama~2.
Specifically, we explore different expert construction methods and continual training strategies to obtain decent models under limited training budgets.
Empirically, \llamoe-3.5B significantly outperforms open-source models with similar activation parameters, such as Sheared-LLaMA-2.7B and Open-LLaMA-3.0B.
Meanwhile, \llamoe-3.0B achieves similar performance with Open-LLaMA-3B with less activated parameters.

From the ablation studies, we find the optimized static data sampling weights could achieve better results, and further data filtering on low-fluency texts also brings extra performance gain.
\llamoe models also show the expert specialization phenomenon, where each expert has domain preferences.
Based on this preference, we explore the expert selection similarities across pre-training datasets and downstream task datasets.

\bibliography{custom}
\bibliographystyle{acl_natbib}

\end{document}